\documentclass[letterpaper, 10 pt, conference]{ieeeconf}
\IEEEoverridecommandlockouts                              % This command is only needed if
\usepackage[T1]{fontenc}
\usepackage[noadjust]{cite}

\usepackage{filecontents}
\usepackage{amsmath}
\DeclareMathSymbol{:}{\mathord}{operators}{"3A}  %to reduce space around colons in math mode. See https://tex.stackexchange.com/questions/432778/reduce-spacing-around-colons-in-math-mode. 
\usepackage{yhmath} %note this package will generate type 3 fonts unless you have 1.6 installed
\usepackage{mathtools} % for rotation matrix e.g. \prescript{4}{0}{R} and for summation
\usepackage{amssymb} %for trianleq symbol
\usepackage{xfrac} % needed for diagonal fraction \sfrac{a}{b} command
\usepackage{sidecap}
\usepackage{graphicx}
\usepackage{subcaption}
\usepackage{multirow} 
\usepackage{fancyhdr}
\usepackage{leftidx} %added by Jared
\usepackage{cancel} %added by Jared
\usepackage{caption}
\usepackage[super]{nth} %to type 1st,2nd,3rd,... , use the command \nth {<number>}
\usepackage[normalem]{ulem} % to strike out
\usepackage{color}
\usepackage{xcolor} %for coloring text
\usepackage{booktabs} %nice tables
\usepackage{siunitx} %e.g. \ang{45} to show 90 degree
\usepackage{booktabs} %nice tables
\usepackage{colortbl} % to color tables
\usepackage{textcomp} %for a nice registered trademark, e.g. MATLAB\textsuperscript{\textregistered}
\usepackage{tikz}
\usepackage[noadjust]{cite} %to group citations
\usepackage{xcolor}
\usepackage{wasysym}  % for diameter symbol e.g. \diameter
\usepackage{arydshln}
\newcommand{\RNum}[1]{\uppercase\expandafter{\romannumeral #1\relax}} %Roman numerals

\newcommand{\realfield}[1]{\hbox{I \kern -.5em R}^{#1}}
\newcommand {\mb}[1]{\mathbf{#1}}
\newcommand {\bs}[1]{\boldsymbol{#1}}

\newcommand{\Rot}[2]{{^{#1}\mathbf{R}}_{#2}}  %example: \Rot{0}{1}
\newcommand{\T}{^{\mathrm{T}}}  %shortcut for transpose
\newcommand*\circled[1]{\tikz[baseline=(char.base)]{\node[circle,minimum size=7pt,draw=black,inner sep=0.5pt](char){\scriptsize #1};}}

\newcommand{\htf}[2]{{}^{#1}\mb{T}_{#2}}

\usepackage{enumerate}
\newif\ifTrackChanges   %define a conditional variable called TrackChanges
\TrackChangesfalse %set TrackChanges to be false. %Uncomment to make a clean version (you have to comment the line above)
\usepackage[textsize=scriptsize, bordercolor=white,backgroundcolor=gray!30,linecolor=black,colorinlistoftodos]{todonotes}  %for adding comments. See http://www.texample.net/tikz/examples/todo-notes/.
\usepackage[normalem]{ulem} % added for strikethrough text use   \sout{text you want to strike through
\usepackage{soul} %added for \texthl for the highlight and also for \ul for underline across wrapped text

\usepackage[mathscr]{euscript}
\ifTrackChanges
    \newcommand{\cut}[1]{{\color{lightgray}{#1}}}
    \newcommand{\corr}[1]{{\color{red}{#1}}}   %correction not requested by reviewer
    \newcommand{\corrlab}[2]{{[\colorbox{yellow}{#1}]~}{\color{red}{#2}}} %correction with a label in response to a reviewer
    \newcommand{\cusst}[1]{{\st{#1}}}  % struck out test 
\else
    \newcommand{\cut}[1]{}
    
    \newcommand{\corrlab}[2]{#2}
    \newcommand{\corr}[1]{#1}
    \newcommand{\cusst}[1]{} 
\fi
\usepackage{algorithm,algorithmicx,algpseudocode}

\algnewcommand\algorithmicinput{\textbf{Input:}}
\algnewcommand\Input{\item[\algorithmicinput]}

\algnewcommand\algorithmicoutput{\textbf{Output:}}
\algnewcommand\Output{\item[\algorithmicoutput]} 

\graphicspath{{figures/pdf/}}

\newbox\tempbox
\makeatletter
\let\NAT@parse\undefined
\makeatother

\usepackage{hyperref} % hyperref still needs to be put at the end!
\hypersetup{
  colorlinks   = true, %true=Colours links instead of ugly boxes
  urlcolor     = blue, %Colour for external hyperlinks
  linkcolor    = blue, %Colour of internal links
  citecolor   = green %Colour of citations
}

\usepackage{balance}
\title{\LARGE \bf A Feasibility Study of a Soft, Low-Cost, 6-Axis Load Cell for Haptics}
\author{Madison Veliky$^{1}$, Garrison L.H. Johnston$^{1}$, Ahmet Yildiz$^{1}$ and Nabil~Simaan$^{1}$$^{\dag}$% <-this % stops a space
\thanks{$\dag$ Corresponding author}% <-this % stops a space
\thanks{$^{1}$Department of Mechanical Engineering, Vanderbilt University, Nashville, TN 37235, USA
        {\tt\small (madison.a.veliky, garrison.L.johnston, ahmet.yildiz, nabil.simaan) @vanderbilt.edu}}%
\thanks{This work was funded by a grant from the Wellcome Leap Trust's SAVE program and Vanderbilt University internal funds.}%
}

\begin{document}
\maketitle

\thispagestyle{empty}  %to suppress the page number of first page
%\pagestyle{plain}

%\thispagestyle{empty}

% Uncomment this section to add header and footer with copyright
\thispagestyle{fancy}
\fancyhf{}
\renewcommand{\headrulewidth}{0pt}
%^^^^^^^^^^^^^^^^^^^^^^^^^^^^^^^^^^^^^^^^^^^^^
%% THE BELOW TWO COMMANDS ARE FOR ARXIV. DELETE/COMMENT FOR FINAL SUBMISSION TO IEEE
%
\lhead{2024  IEEE/RSJ International Conference on Intelligent Robots and Systems (IROS 2024). Accepted Version. }
\rfoot{\centering \scriptsize \copyright 2024 IEEE. Personal use of this material is permitted. Permission from IEEE must be obtained for all other uses, in any current or future media, including reprinting/republishing this material for advertising or promotional purposes, creating new collective works, for resale or redistribution to servers or lists, or reuse of any copyrighted component of this work in other works.}
%=============================================
\pagestyle{empty}

%----------------------------------------------------------------------------------------------------------%
%-----------------------------------------------ABSTRACT---------------------------------------------------%
%----------------------------------------------------------------------------------------------------------%
%----------------------------------------------------------------------------------------------------------%
%----------------------------------------------ABSTRACT----------------------------------------------------%
%----------------------------------------------------------------------------------------------------------%
% \begin{abstract}

\begin{abstract}
 Haptic devices have shown to be valuable in supplementing surgical training, especially when providing haptic feedback based on user performance metrics such as wrench applied by the user on the tool. However, current 6-axis force/torque sensors are prohibitively expensive. This paper presents the design and calibration of a low-cost, six-axis force/torque sensor specially designed for laparoscopic haptic training applications. The proposed design uses Hall-effect sensors to measure the change in the position of magnets embedded in a silicone layer that results from an applied wrench to the device. Preliminary experimental validation demonstrates that these sensors can achieve an accuracy of 0.45 N and 0.014 Nm,  and a theoretical XY range of $\pm$50N, Z range of $\pm$20N, and torque range of $\pm$0.2Nm. This study indicates that the proposed low-cost 6-axis force/torque sensor can accurately measure user force and provide useful feedback during laparoscopic training on a haptic device.  

\end{abstract}

%----------------------------------------------------------------------------------------------------------%
%----------------------------------------------KEYWORDS----------------------------------------------------%
%----------------------------------------------------------------------------------------------------------%

%----------------------------------------------------------------------------------------------------------%
%-----------------------------------------------CONTENTS---------------------------------------------------%
%----------------------------------------------------------------------------------------------------------%
\section{Introduction} \label{sec:intro}
\par Haptic feedback has been incorporated into a wide range of devices for various applications such as perception augmentation, enforcing safety barriers, and applying assistive control behaviors (virtual fixtures). Accordingly, haptic devices have become increasingly popular for surgical training as they can provide either tactile or force cues on the trainee as guidance/feedback on their performance. To give effective feedback, a haptic device should ideally be able to discern the motions and forces applied to the tool. However, many haptic trainers neglect force measurement due to a lack of affordable and easily integrable force/torque (F/T) sensor options. This paper seeks to resolve this issue with the design of a \corr{soft,} low-cost, 6-axis F/T sensor designed specifically for integration within a laparoscopic haptic trainer for skill assessment and surgical training. 
\par The majority of commercially available 6-axis F/T sensors are comprised of miniature strain gauges applied to precise micro-machined metal flexures \cite{Templeman2020,Billeschou2021,Ahmad2021}. This method generates very accurate sensors, but comes at a very high cost of fabrication and specialized circuitry for amplification and signal conditioning. The typical price range for this type of design is several thousands of U.S. dollars. This proves to be a major limitation for incorporation of force sensing in haptic devices. Pursuing a low-cost design will improve the feasibility of exploiting force data for feedback in surgical assessment and increase access to such haptic trainers. The cost of fabrication of these F/T sensors may be reduced by 3-D printing the flexures into the sensor (e.g. \cite{Yao2016}), but the cost and complexity of signal amplification and conditioning remains contra-indicated to the simplicity we seek per our application domain specifications.  

\par Other low-cost load cells have been proposed in the literature. For example, capacitance-based force sensors are significantly less expensive, but also tend to be less accurate than other force sensing methods \cite{Kim2017,He2021,Rahman2022}. Another approach attempts to create flexures with optical sensors \cite{Tar2011,Gafford2016,Al-Mai2018,Noh2019,Hendrich2020,Li2022} to reduce cost. However, these sensors are subject to reliability issues based on the accuracy of the 3D printer used. Another method for creating F/T sensors is to embed magnets inside flexible material and measure the change in magnetic field using a Hall-effect sensor. This is an attractive option due to their simple electronics and construction. Abah et al. \cite{Abah2019,Abah2022} have constructed single axis force sensors for integration into continuum robots using Hall-effect sensors. In \cite{JunJiang2013}, the authors theorize a 2-axis force sensor from Hall-effect sensors and flexures. Gomez et al. \cite{Gomez2015} use Hall-effect sensors to measure the center of pressure for a humanoid robot's foot. The sensors proposed in \cite{Tomo2016,Paulino2017,Harber2020,MoosaviNasab2022} all use Hall-effect sensors to create a 3-axis force sensor. Lastly, Nie and Sup \cite{Nie2017} present a four axis, soft F/T sensor for robotic applications. Despite their potential benefits, to the best of our knowledge, Hall-effect sensors have not been used to create a 6-axis F/T sensor.
\par In this paper, we detail the design, modeling, and calibration of the first soft, 6-axis F/T sensor that employs Hall-effect sensors for surgical applications. The closest work to ours is Harber et al. \cite{Harber2020} where the authors propose integrating their sensor into the Da Vinci Research Kit for tumor palpation. However, their sensor only measures forces (i.e. 3-axis) and their sensor is specifically designed for palpation and not for general laparoscopic training.  
\par For the chosen application domain of haptic devices for soft-tissue surgery, the need for low-cost and single-use devices outweighs the need for high-stiffness and high dynamic bandwidth due to slow interaction with soft tissue. Driven by these observations, we have set forth with an exploratory design of a six-axis F/T sensor that meets these requirements. 
\par \cut{Another important question to consider is how can forces at the tip of a surgical instrument be estimated if mounted by two of the proposed force sensors? Surgical training and skill assessment are more concerned with the force applied by the tool tip on its environment than the forces on the shaft of the tool. Therefore, investigating this question is important since it allows a low-cost and unobtrusive way of estimating loads at the tool tip of interchangeable instruments when inserted into our future haptic trainer. To answer this question, we consider two force sensors mounted on the tool shaft at a known distance from each other.}
\par The contribution of this paper is twofold. First, we present the novel design of a 6-axis load cell including a discussion of the fabrication and calibration process. The second contribution is a model of the uncertainty propagation for the sensor\cut{ and an approach for indirect force sensing at the tool-tip of a laparoscopic instrument}. Since this is an early feasibility study, we exclude the effects of geometric and material uncertainties and consider only the effects of low-level measurement uncertainties. \corrlab{R4}{We also reserve considerations of drift, hysteresis, and repeatability for a more detailed investigation once these uncertainties are resolved.}
\par The paper first presents the sensor design including the details of the sensing mechanism, mechanical design, electrical design, and the software configuration parameters selected. Next, an overview of the force sensing model is discussed. We then describe the method for determining the magnetic field to position mapping. The method of calibration and validation of the force sensor follows. We conclude with a discussion of the proposed sensor's characteristics and ability to serve our application.

%========================================================
\section{Sensor Design}\label{sec:sensor_design}
\subsection{Sensing Mechanism} \label{sec:sensing_mechanism}
\par The mechanism of sensing force is based on Hooke's Law, which gives a direct relationship between force applied on a spring and its deformation. Silicone has highly elastic properties \cut{which enable it to approximate an ideal spring.} \corrlab{R4-8}{and typically follows a non-linear deformation model. However, under small deformations, the relationship between force and deformation can be approximated as linear for the purposes of this initial feasibility study.} In order to measure deformation, small 1/16" diameter, 1/32" height magnets (K\&J Magnetics, D101-N52) are placed on the surface of a layer of silicone and Hall-effect sensors/magnetometers (Melexis, MLX90393) are placed on the opposite surface to measure magnetic field strength. As the silicone is compressed, the relative position between the magnets and the magnetometers changes which is captured by the change in magnetic flux density. We assume a linear mapping between position and magnetic field strength, as given by:
\begin{equation}
    \label{eq:map}
    \mathbf{p} = \mathbf{M} \mathbf{b} + \mathbf{o}
\end{equation}

\noindent where $\mathbf{p}$ is the position of the magnet with respect to the Hall-effect sensor, $\mathbf{M}$ is a diagonal matrix containing the change in position for a change in flux for each axis, $\mathbf{b}$ is the flux density, and $\mathbf{o}$ is the sensor offset. 

\par Any one-dimensional force applied to an isolated magnet, silicone, Hall-effect sensor system can be calculated using Hooke's Law:

\begin{equation} \label{eq:mechanism}
    F_i = k (m b_i + o)
\end{equation}
where $F_i$ is the force applied to the sensor, $k$ is the spring constant of the silicone, $m$ is the constant that relates the distance of the magnet to magnetic field strength, $b$ is magnetic field strength, and $o$ is some offset. Extrapolating to six dimensions from this one-dimensional case is explained in detail in Section \ref{sensing_model}. 

\subsection{Mechanical Design} \label{sec:mechanical_design}
\par
\begin{figure}[tbp]
        \centering
        % trim={<left> <lower> <right> <upper>}
        \includegraphics[width=\columnwidth]{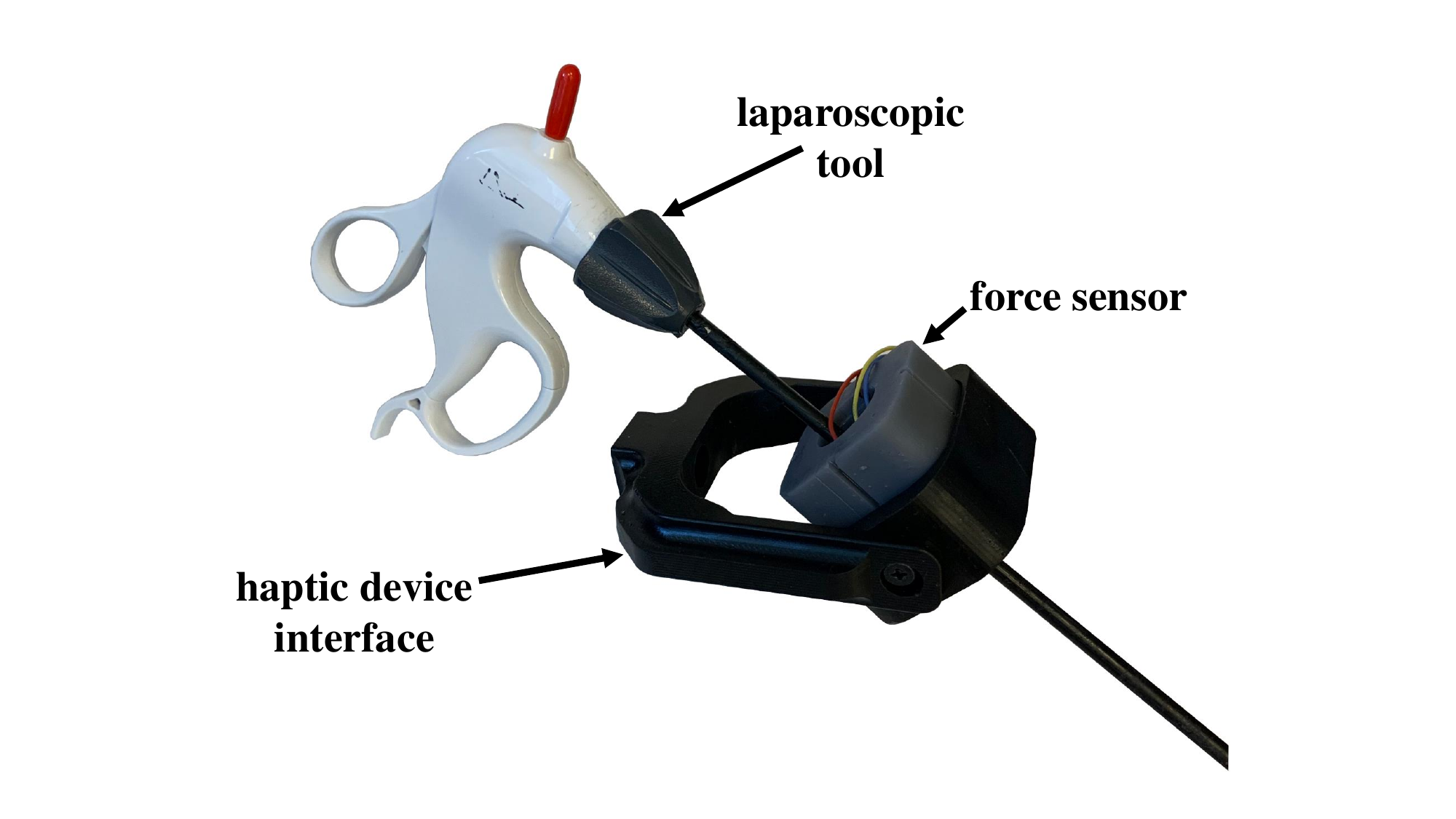}
        \caption{Integration of the proposed force sensor with a laparoscopic tool and haptic device.}
        \label{fig:tool_integration}
\end{figure}

\begin{figure}[tbp]
        \centering
        % trim={<left> <lower> <right> <upper>}
        \includegraphics[width=0.85\columnwidth]{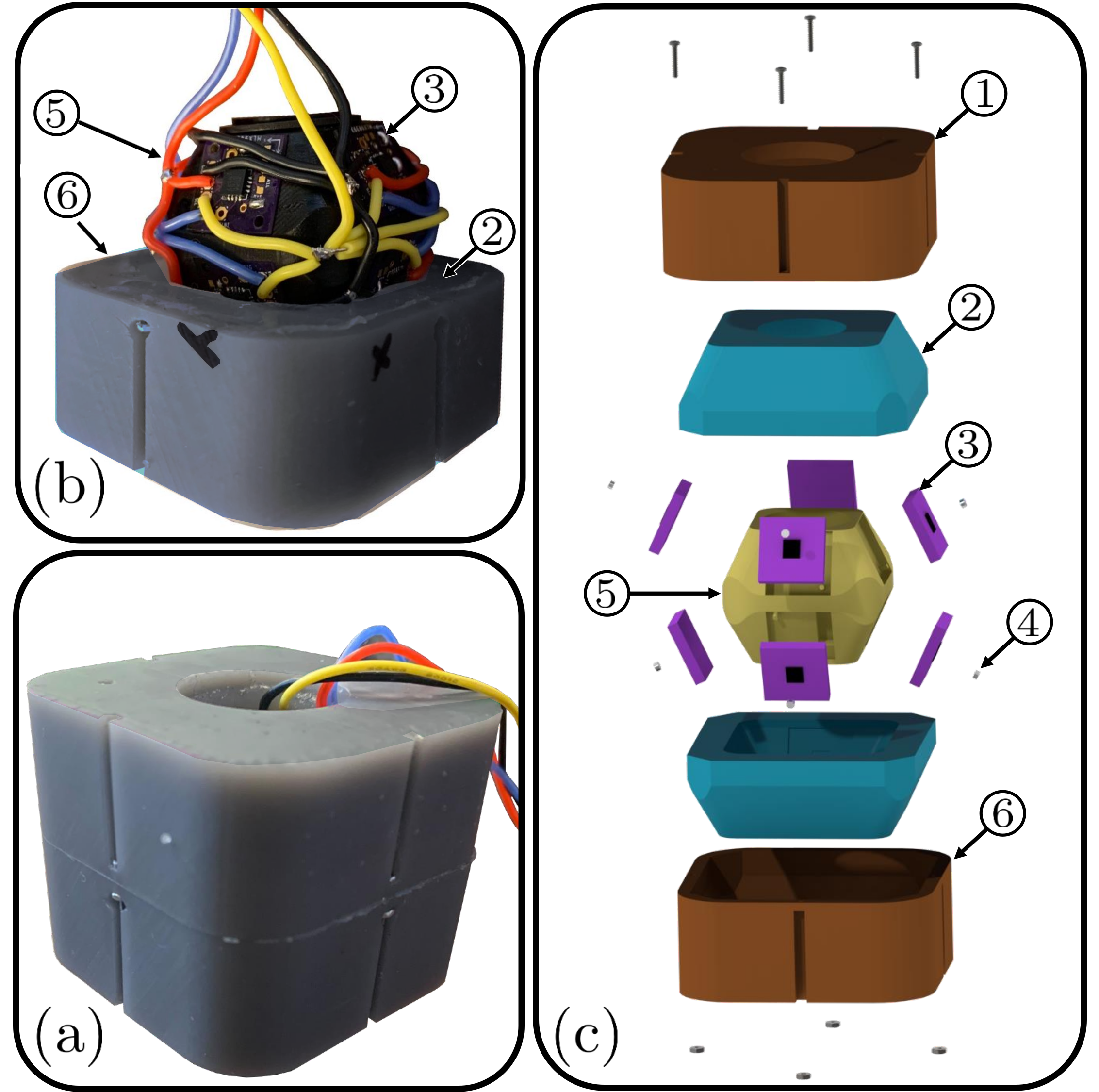}
        \caption{(a) Picture of the assembled force/torque sensor. (b) Picture of the sensor with the top cover and top silicone layer removed. (c) Exploded view of sensor: \protect\circled{1} top outer shell, \protect\circled{2} silicone layer, \protect\circled{3} MLX90393 Magnetometer, \protect\circled{4} K\&J Magnetics D101- N52 magnet, \protect\circled{5} center piece, \protect\circled{6} bottom outer shell.}
        \label{fig:exploded_view}
\end{figure}

 \corrlab{R4-5}{While the sensor can also be used as a general 6-axis F/T sensor, it was designed with }the target application of measuring forces and torques applied at the tip of surgical tools in a haptic training device. The sensor is designed such that it can be easily mounted to laparoscopic instruments by passing the shaft of the tools through the center of the force sensor as shown in Fig. \ref{fig:tool_integration}. The sensor itself has three main components: a center piece with attached Hall-effect sensors, a layer of silicone, and an outer shell to house the magnets (Fig. \ref{fig:exploded_view}). Both the center piece and the outer shell are 3D printed out of Phrozen Rock-Black Stiff resin (Phrozen, Sonic Mega 8K S). The center piece serves to rigidly attach the magnetometers to the tool shaft whereas the magnets are fixed to the haptic device. The layer of silicone \corrlab{R4-7}{is molded with Eco-Flex 00-30 (Smooth-On) which} has a shore hardness of 00-30. The layer thickness is 6mm between the center piece and the outer shell. This allows the sensors to move with respect to the magnets proportional to the amount of force applied by the user on the tool shaft. The materials and associated cost estimation are provided in table \ref{tab:cost}. 

\begin{table}[h]
    \caption{Materials and cost estimation. }\label{tab:cost}
    \centering
    \begin{tabular}{|c|c|c|c|}
        \hline
        \textbf{Material} & \textbf{Price per Unit} & \textbf{Amount} & \textbf{Price}\\ \hline
        Resin & \$66.19/kg & 0.0533kg & \$3.53\\
        \hline
        Silicone & \$1.34/oz & 1oz & \$1.34\\
        \hline
        Magnets & \$0.14/unit & 8 units & \$1.12\\
        \hline
        Magnetometers & \$1.95/unit & 8 units & \$15.60\\
        \hline
        PCBs & \$0.33/unit & 8 units & \$2.64 \\
        \hline
        Wires & \$0.08/ft & 4ft & \$0.32\\
        \hline 
        Screws & \$1.60/unit & 4 units & \$6.4\\
        \hline
        Nuts & \$0.45/unit & 4 units & \$1.74\\
        \hline
         \multicolumn{3}{|c|}{\textbf{Total}} & \textbf{\$32.74} \\ \hline
    \end{tabular}
\end{table}

\par
\begin{figure}[tbp]
        \centering
        % trim={<left> <lower> <right> <upper>}
        \includegraphics[width=0.75\columnwidth]{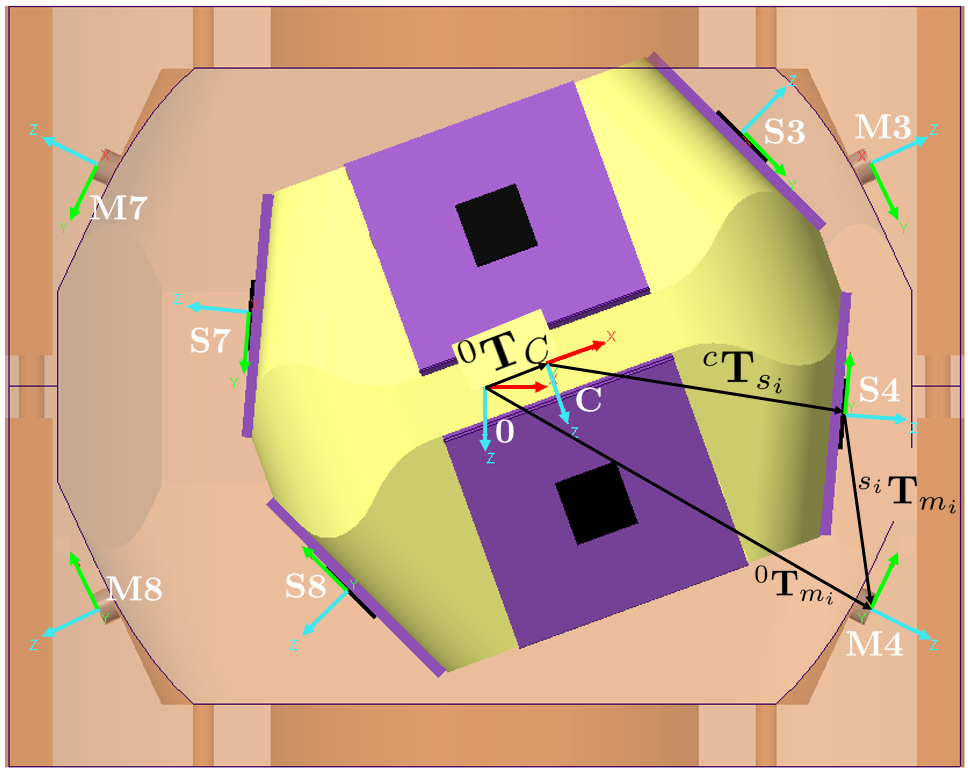}
        \caption{Frame assignments for the nominal center frame \{0\}, the deformed center frame \{C\}, the sensor frames \{$S_i$\}, and the magnet frames \{$M_i$\}. \corrlab{R4-10}{See Multimedia Extension 1 for a visualization of how the center piece deforms under a load.} }
        \label{fig:frame_assignment}
\end{figure}

 To capture the forces/torques in six degrees of freedom, the sensor has eight magnetometers and eight corresponding magnets arrayed radially around the center of the tool. The Hall-effect sensors sit at an angle of 25$^\circ$ from the vertical axis of the tool shaft. The magnets are offset by 6mm from the surface of the sensors. To encapsulate the positional deformation that results from a force applied to the sensor, the position of each magnet with respect to its magnetometer must be combined to give the deformation twist of the center piece as a whole. \corrlab{R4-9}{Eight sensors were chosen to provide redundancy that can be leveraged to minimize error in the computation of the deformation twist.} This is explained in more detail in section \ref{sensing_model}. As a basis for that computation, it \corr{is necessary} to define a coordinate system for each magnetometer and magnet with respect to the base frame of the force sensor in the nominal case (no force applied) (Fig. \ref{fig:frame_assignment}). The base frame is placed at the center of the device is defined with the z-axis along the tool shaft in the direction of the tool tip. The Hall-effect sensor frames are assigned by the chip itself and have the z-axis pointing outward from the face of the chip, the y-axis pointing along the length of the tool shaft, and the x-axis along the horizontal surface of the inner ring. The frames of the magnets have the exact same orientation as their corresponding magnetometers, just translated 6mm in the positive Z direction.

\subsection{Electrical Design} \label{sec:electrical_design}
\par
\begin{figure}[tbp]
        \centering
        % trim={<left> <lower> <right> <upper>}
        \includegraphics[ width=0.95\columnwidth]{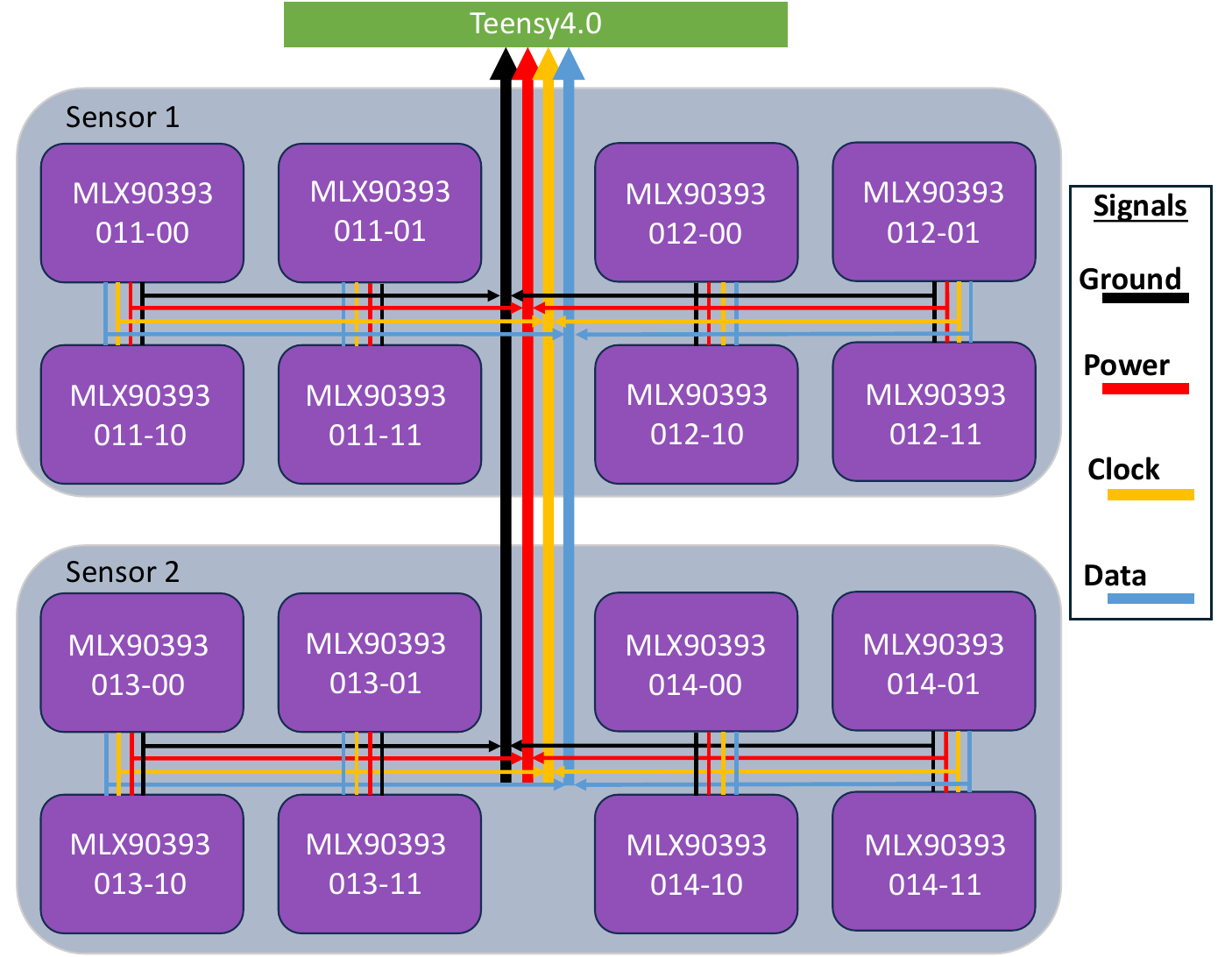}
        \caption{I2C wiring diagram. There exist 16 unique I2C addresses for the MLX90393 magnetometer which allows up to two F/T sensors on the same I2C bus. }
        \label{fig:electrical}
\end{figure}

 Measurements from the Hall-effect sensors are sent via I2C communication protocol to a Teensy 4.0 microcontroller. There are four \cut{different} versions of the MLX90393 magnetometer chip which can each accommodate four user-specified addresses, so in total, 16 MLX90393 magnetometers can fit on a single I2C bus (Fig. \ref{fig:electrical}). Since the current sensor design uses eight Hall-effect sensors, two of our force sensing devices can be connected on a single I2C bus, which will facilitate future integration into a haptic device. Each magnetometer is soldered to a custom printed circuit board (PCB) which interfaces between each chip and assigns the I2C address for each magnetometer.

\subsection{Chip Configuration}
\par The MLX90393 magnetometer offers several parameters that can be adjusted by the user to determine the quantities, resolution, and frequency of magnetic field measurements. The Hall-effect sensor is capable of detecting changes in magnetic field in the X, Y, and Z direction as well as the internal temperature of the chip. For this application, measurements in all three dimensions are considered and temperature is ignored. 

\par Information from the magnet supplier indicates that the magnetic field strength at the surface of the magnet is 6053 gauss (605.3 mT). Therefore, the sensor must be capable of capturing the range 0-605300$\mu$T. The magnetometer can report 16 bits per measurement which means that the ideal resolution should be no less than 9.236 $\mu$T/LSB in the Z-direction. The best option for resolution offered by the MLX90393 is 9.680 $\mu$T/LSB along the Z-axis. The corresponding resolution for the field strength in the X and Y directions are both 6.009 $\mu$T/LSB. 

\par The maximum frequency of data acquisition is limited by the time it takes to perform a measurement. This is determined by the amount of time needed to prepare the magnetometer for a measurement and how many dimensions must be measured. For our application, the preparation time is 839$\mu$s and the measurement time for a single axis is 835$\mu$s. Therefore the minimum period for measuring from three axes is 3.34ms. A 10ms period is chosen to give a sampling frequency of 100Hz, which is far below the threshold for human tactile perception of latency \cite{Doxon2013}.

\section{Force Estimation}\label{sensing_model}
% \begin{enumerate}
%     \item Sensor $i$ senses magnetic flux density $\mb{b}_i = [b_x,~b_y,~b_z]\T$ in Teslas with respect its own local frame $\{S_i\}$.
%     \item Using the experiments given in section ??? we calibrated the position of the magnet ${}^{s_i}\mb{p}_{m_i}$ as a function of $\mb{b}_i$ 
%     \item Using Arun's method \cite{Arun1987_pose_est}, we can estimate the deformation twist of the center piece.
%     \item For simplicity, we then fit a linear model to estimate the deflection twist $\Delta\bs{\xi}$ from $\mb{b} = [\mb{b}_1\T,\dots,~\mb{b}_8\T]\T$. $\Delta\bs{\xi} = \mb{A}\mb{b}$
%     \item We can now estimate the applied wrench from ${}^0\mb{w} = \mb{K}\Delta\bs{\xi}$. 
%     \item We estimate the stiffness matrix using ...
%     \item The final force sensing model is then $\mb{w} = \mb{K}\mb{A}\mb{b}$
%     \item $\Delta\mb{w} = \mb{K}\mb{A}(\Delta\mb{b})$
% \end{enumerate}

As previously mentioned, when a wrench is applied to the sensor's center piece (Fig. \ref{fig:exploded_view}-\circled{5}), the center deflects inside of the silicone layer (Fig. \ref{fig:exploded_view}-\circled{2}). In this section, we describe a method for estimating the wrench applied to the sensor's center piece and the pose of the center piece using the measurements from the MLX90393 sensors (Fig. \ref{fig:exploded_view}-\circled{3}).
\par The $i^\text{th}$ MLX90393 sensor measures the magnetic flux density $\mb{b}_i = [b_x,~b_y,~b_z]\T$ in Teslas in its own local frame $\{S_i\}$. As described in section \ref{sec:mapping}, we experimentally calibrated the position of the $i^\text{th}$ magnet ${}^{s_i}\mb{p}_{m_i}$ as a function of $\mb{b}_i$ in $\{S_i\}$. Next, we describe how we use the measurements of ${}^{s_i}\mb{p}_{m_i}$ to estimate the pose of the center piece using the Arun's method for point-cloud registration \cite{Arun1987_pose_est}.   
\subsection{Center Piece Pose Estimation} \label{sec:center_pose_est}
\par As shown in Fig. \ref{fig:frame_assignment}, we assign a frame $\{C\}$ that is fixed to the sensor center piece. When no wrench is applied to the sensor, this frame will be aligned with the original center frame (i.e., $\htf{0}{c} = \mb{I}_4$). When deflected, $\{C\}$ exhibits a positional and/or orientational offset from the world frame (i.e., $\htf{0}{c} \neq \mb{I}_4$). To solve for $\htf{0}{c}$, we use the following loop closure equation:
\begin{equation}\label{eq:loop_closure_1}
    {}^0\mb{p}_{m_i} = {}^0\mb{p}_{c} + \Rot{0}{c}{}^c\mb{p}_{s_i} + \Rot{0}{c}\Rot{c}{s_i}{}^{s_i}\mb{p}_{m_i}
\end{equation}
In this equation, ${}^0\mb{p}_{c}\in\realfield{3}$ and $\Rot{0}{c}\in SO(3)$ are the translational and orientational components of $\htf{0}{c}$, respectively. ${}^0\mb{p}_{m_i}\in\realfield{3}$ is the position of the $i^\text{th}$ magnet in world frame. This value is a constant that is known from the sensor geometry. Additionally, ${}^c\mb{p}_{s_i}\in\realfield{3}$ is the position of the $i^\text{th}$ Hall-effect sensor frame $\{S_i\}$ relative to $\{C\}$ which is also a constant value and ${}^{s_i}\mb{p}_{m_i}\in\realfield{3}$ is the position of the $i^\text{th}$ magnet as sensed by the $i^\text{th}$ Hall-effect sensor. This equation can be simplified to: 
\begin{equation}\label{eq:loop_closure_2}
    {}^0\mb{p}_{m_i} = {}^0\mb{p}_{c} + \Rot{0}{c}{}^{c}\mb{p}_{m_i}
\end{equation}
\par Using this equation and information from all 8 Hall-effect sensors, we can estimate ${}^0\mb{p}_{c}$ and $\Rot{0}{c}$ using a least-squares approach:     
\begin{equation}
        \min_{\Rot{0}{c},{}^0\mb{p}_{c}} \quad \sum_{i=1}^8\|{}^0\mb{p}_{m_i} - \left(\Rot{0}{c}{}^c\mb{p}_{m_i} + {}^0\mb{p}_{c}\right)\|^2 
\end{equation}
To solve this equation, we rely on the singular value decomposition (SVD) based approach first proposed by Arun et al. \cite{Arun1987_pose_est}. In this method, we first find the centroid of the position measurements:
\begin{equation}
    {}^0\overline{\mb{p}}_{m} = \frac{1}{8}\sum_{i = 1}^8{}^0\mb{p}_{m_i},\quad  {}^c\overline{\mb{p}}_{m} = \frac{1}{8}\sum_{i = 1}^8{}^c\mb{p}_{m_i}
\end{equation}
Using these centroids, we then calculate the following $3\times3$ matrix:
\begin{equation}
    \mb{H} = \sum_{i=1}^8\left({}^c\mb{p}_{m_i} - {}^c\overline{\mb{p}}_{m}\right)\left({}^0\mb{p}_{m_i} - {}^0\overline{\mb{p}}_{m}\right)\T
\end{equation}
Next, we take the SVD of $\mb{H}$:
\begin{equation}
    \mb{H} = \mb{U}\bs{\Sigma}\mb{V}\T
\end{equation}
The rotation matrix can now be estimated using:
\begin{equation}\label{eq:center_piece_rot}
    \Rot{0}{c} = \mb{V}\mb{U}\T
\end{equation}
Once the rotation matrix is estimated, the translation can be estimated using:
\begin{equation}\label{eq:center_piece_pos}
   {}^0\mb{p}_{c} = {}^0\overline{\mb{p}}_{m} - \Rot{0}{c}{}^c\overline{\mb{p}}_{m}
\end{equation}
Using \eqref{eq:center_piece_rot} and \eqref{eq:center_piece_pos}, we can write the pose of the center piece with respect to the sensor world frame in the form of an $SE(3)$ transformation matrix $\htf{0}{c}$. 
%
% \begin{equation}
%     \htf{0}{c} = \begin{bmatrix}
%         \Rot{0}{c} & {}^0\mb{p}_{c} \\ \mb{0} & 1
%     \end{bmatrix} \in SE(3)
% \end{equation}
%
\subsection{Deflection Twist} \label{sec:deflection_twist}
For our force sensing model, we want to express the deflection of the center piece in terms of the twist $\Delta\bs{\xi}\in\realfield{6}$. We can extract the deflection twist from $\htf{0}{c}$ using: 
\begin{equation}
    \Delta\bs{\xi} = \log\left(\htf{0}{c}\right)^\vee\in\realfield{6}
\end{equation}
Where $\log(\cdot):~SE(3)\rightarrow se(3)$ is the matrix logarithm that takes an $SE(3)$ transformation matrix and returns a matrix $se(3)$ twist and $(\cdot)^\vee: ~se(3)\rightarrow\realfield{6}$ takes a matrix $se(3)$ twist and returns a twist in vector form \cite{Murray1994}.
\subsection{Linear Estimation of Deflection Twists} \label{sec:lin_def_twist}
For simplicity, we use the estimation method described in the previous section to fit a matrix $\mb{A}\in\realfield{6\times24}$ that can be used estimate the deflection twist $\Delta\bs{\xi}$ from a stacked vector of magnetic flux densities from all eight sensors $\widehat{\mb{b}} = [\mb{b}_1\T,\dots,~\mb{b}_8\T]\T\in\realfield{24}$:  
\begin{equation}\label{eq:A}
    \Delta\bs{\xi} = \mb{A}\widehat{\mb{b}} 
\end{equation}
To find this matrix, we experimentally measure $n$ different magnetic flux density vectors using the method described in section \ref{sec:calibration} and put them in the columns of a matrix $\mb{B} = \left[\widehat{\mb{b}}_1, \dots, \widehat{\mb{b}}_n \right]\in\realfield{24\times n}$. Using the method described in section \ref{sec:center_pose_est}, we then find the $n$ deflection twists that correspond to the columns of $\mb{B}$ and put them in the columns of the matrix $\bs{\Xi} = \left[\Delta\bs{\xi}_1, \dots,  \Delta\bs{\xi}_n \right]\in\realfield{6\times n}$. Using $\bs{\Xi}$ and $\mb{B}$, \eqref{eq:A} can be rewritten using all $n$ measurements of $\widehat{\mb{b}}$: 
\begin{equation}
    \bs{\Xi} = \mb{A}\mb{B}
\end{equation}
Using this equation, we can find $\mb{A}$ using:
\begin{equation} \label{eq:solve_A}
    \mb{A} =  \bs{\Xi}\mb{B}^+
\end{equation}
Where $\mb{B}^+$ is the Moore-Penrose pseudoinverse of $\mb{B}$.

\subsection{Force Sensing}
Using Hooke's law, the deflection twist $\Delta\bs{\xi}$ can be used to find the wrench $\mb{w}\in\realfield{6}$ applied to the sensor center piece using the stiffness matrix $\mb{K}\in\realfield{6\times6}$: 
\begin{equation}
     \mb{w} = \mb{K}\Delta\bs{\xi}
\end{equation}
By substituting \eqref{eq:A} in for $\Delta\bs{\xi}$, we can find $\mb{w}$ using the magnetic flux density measurements $\widehat{\mb{b}}$:
\begin{equation}\label{eq:wrench}
    \mb{w} = \mb{K}\mb{A}\widehat{\mb{b}} 
\end{equation}
In the experiments described in section \ref{sec:calibration}, we also collected $n$ measurements of the applied wrench. Similar to section \ref{sec:lin_def_twist}, we create a matrix whose columns are the wrench measurements $\mb{W} = \left[\mb{w}_1, \dots, \mb{w}_n \right]\in\realfield{6\times n}$.
\begin{equation}
    \mb{W} = \mb{K}\mb{A}\mb{B}
\end{equation}
We can now find the stiffness matrix using:
\begin{equation} \label{eq:solve_stiffness}
   \mb{K} = \mb{W}\left(\mb{A}\mb{B}\right)^+
\end{equation}

\cut{\subsection{Force Estimation at Tool Tip}
\begin{figure}[h]
    \centering
    \includegraphics[width=0.95\columnwidth]{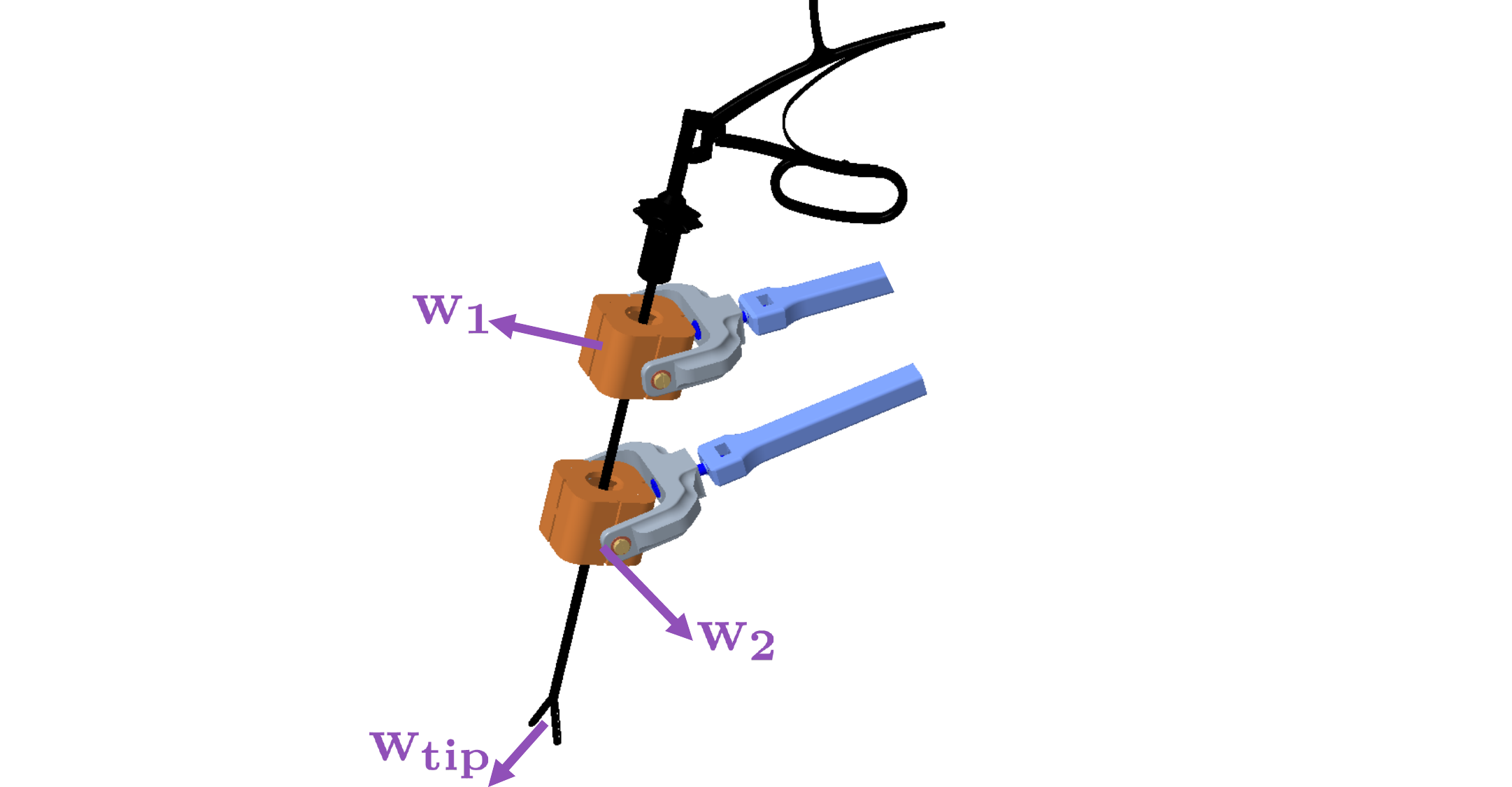}
    \caption{\cut{Estimated wrench at the tip of the tool based on sensed wrench at two points along the tool.} }
    \label{fig:tool}
\end{figure}
As shown in Fig. \ref{fig:tool}, in our laparoscopic surgery trainer, the laparoscopic tool is held using two of our soft force/torque sensors. In this section, we will describe how the wrench measurements from the two sensors can be used to determine the wrench applied to the tip of the laparoscopic tool $\mb{w}_{tip} = [\mb{f}_{tip}\T,\mb{m}_{tip}\T]\T$. The sensors are assembled such that the orientation of their frames are parallel with the frame at the tip of the tool. Because the frames are parallel, the force at the tip is the sum of the forces sensed by the two sensors:
\begin{equation}\label{eq:f_tip}
    \mb{f}_{tip} = \mb{f}_1 + \mb{f}_2
\end{equation}
here $\mb{f}_1$ is the force sensed by the first sensor as shown in Fig. \ref{fig:tool} and $\mb{f}_2$ is the force sensed by the second sensor as shown in Fig. \ref{fig:tool}. The moment at the tool tip is given by:
\begin{equation}\label{eq:m_tip}
    \mb{m}_{tip} = \mb{m}_1 + \mb{m}_2 + \mb{p}_{1/{tip}}\times\mb{f}_1 + \mb{p}_{2/{tip}}\times\mb{f}_2 
\end{equation}
Where $\mb{m}_1$ is the moment sensed by the first sensor, $\mb{m}_2$ is the moment sensed by the second sensor, $\mb{p}_{1/{tip}}$ is the position vector from the tool tip to the center frame of sensor 1, and $\mb{p}_{2/{tip}}$ is the position vector from the tool tip to the center frame of sensor 2. The position vectors $\mb{p}_{1/{tip}}$ and $\mb{p}_{2/{tip}}$ are written in the tip frame that is parallel with the frames of both sensors.
\par Equations \eqref{eq:f_tip} and \eqref{eq:m_tip} can be combined to find the $\mb{w}_{tip}$ in terms of the wrench sensed by sensor 1 $\mb{w}_1$ and the wrench sensed by sensor 2 $\mb{w}_2$:
\begin{equation}
    \mb{w}_{tip} = \underbrace{\begin{bmatrix}
        \mb{I} & \mb{0} \\ \mb{p}_{1/{tip}}^\wedge & \mb{I} \end{bmatrix}}_{\mb{Ad}_1}\mb{w}_1 
    + \underbrace{\begin{bmatrix} \mb{I} & \mb{0} \\ \mb{p}_{2/{tip}}^\wedge & \mb{I}   \end{bmatrix}}_{\mb{Ad}_2}\mb{w}_2
\end{equation}
In this equation, the operator $(\cdot)^\wedge:~\realfield{3} \rightarrow so(3)$ transforms a vector to $3\times3$ skew-symmetric cross product form. To write the wrench at the tip in terms of the measured magnetic flux density, we can substitute \eqref{eq:wrench} in for $\mb{w}_1$ and $\mb{w}_2$:
\begin{equation}
    \mb{w}_{tip} = \mb{Ad}_1\mb{K}_1\mb{A}_1\widehat{\mb{b}}_1
    + \mb{Ad}_2\mb{K}_2\mb{A}_2\widehat{\mb{b}}_2
\end{equation}
This equation can be rewritten as:
\begin{equation}\label{eq:w_tip}
    \mb{w}_{tip} = \begin{bmatrix}
        \mb{Ad}_1\mb{K}_1\mb{A}_1 & \mb{Ad}_2\mb{K}_2\mb{A}_2
    \end{bmatrix}\begin{bmatrix} \widehat{\mb{b}}_1 \\ \widehat{\mb{b}}_2\end{bmatrix}
\end{equation}
}
\section{Sensitivity and Error Analysis}
In this section, we aim to establish theoretical bounds on the uncertainty in the estimated wrench given the uncertainty in the magnetic flux density. In this initial feasibility study, we consider only the uncertainty due to magnetic flux density measurement and ignore the contribution of sensor geometry \corrlab{linearity assumption}{and material properties of silicone} to the uncertainty in the wrench estimation.  
\par Given \eqref{eq:wrench}, the deviation in the wrench estimation $\delta\mb{w}$ that corresponds to an unmodeled measurement error in magnetic flux density $\delta\widehat{\mb{b}}$ is given simply by:
\begin{equation}
    \delta\mb{w} = \mb{K}\mb{A}\delta\widehat{\mb{b}}
\end{equation}
In many cases, we do not know the exact value of $\delta\widehat{\mb{b}}$. We are therefore more interested in establishing bounds on the norm of $\delta\mb{w}$. Therefore, we write the following inequality:
\begin{equation}
    \|\delta\mb{w}\| \leq \|\mb{K}\mb{A}\|\|\delta\widehat{\mb{b}}\|
\end{equation}
where $\|\mb{K}\mb{A}\|$ is the spectral or 2-norm of $\mb{K}\mb{A}$. Given that the 2-norm of a matrix equals its maximum singular value \cite{strang2006_linear_alg}, this equation can be rewritten as:  
\begin{equation}
    \|\delta\mb{w}\| \leq \sigma_{\max}(\mb{K}\mb{A})\|\delta\widehat{\mb{b}}\|
\end{equation}
Therefore, the amplification of the measurement errors in $\widehat{\mb{b}}$ is bounded by the maximum singular value $\sigma_{\max}(\mb{K}\mb{A})$.  
\par Similarly, in \eqref{eq:w_tip}, the amplification of uncertainties in measurement error in $\widehat{\mb{b}}_1$ and $\widehat{\mb{b}}_2$ are bounded by the maximum singular value of $\begin{bmatrix} \mb{Ad}_1\mb{K}_1\mb{A}_1 & \mb{Ad}_2\mb{K}_2\mb{A}_2 \end{bmatrix}$.     
\section{Experiments} \label{sec:experiments}
\par To verify the sensor design and force sensing model, a number of experiments were performed which are described in the following sections. First, the relationship between individual displacement of the magnets based on magnetic field strength was determined experimentally. Then the sensor was calibrated with a robust combination of forces and torques to populate the stiffness matrix, $\mb{K}$ (\ref{eq:wrench}). Finally, the force sensing model was verified by applying a set of known forces/torques to the sensor and comparing the sensed force/torque to the ground truth. 
\subsection{Position Mapping} \label{sec:mapping}
\par As mentioned in section \ref{sec:sensing_mechanism}, we assume a linear relationship between the measured  magnetic flux density and position of the magnet with respect to the sensor. The MLX90393 magnetometer is able to detect magnetic flux density along three axes (X, Y, and Z). We first verified that the change in magnetic flux for each dimension is sufficiently sensitive to a positional change in that dimension (i.e. a change in the x-direction corresponds to a change in the x-component of flux density). The following experiment was performed to determine the relationship between a change in position and the corresponding change in magnetic flux.   
\par
\begin{figure}[t]
    \centering
    \includegraphics[width=0.3\textwidth]{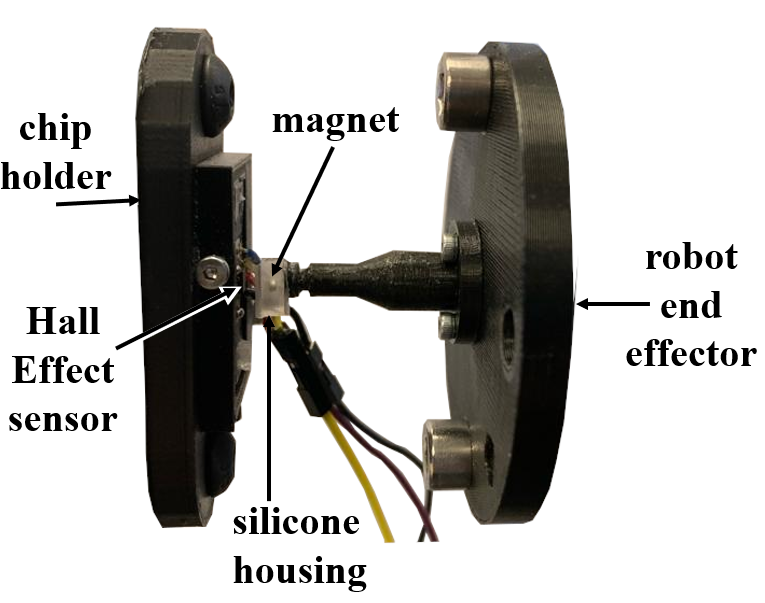}
    \caption{Setup for sensitivity experiments}
    \label{fig:sens_setup}
\end{figure}

 The setup is composed of a motorized linear XYZ-stage robot, a 3D printed mount for the Hall-effect sensor, a magnet embedded in silicone housing and a Teensy4.0 to read the data serially. The data is recorded for post-processing using the serial port logging feature of the open source software puTTy. The magnet is mounted to the end effector of the Cartesian robot which is zeroed by aligning the magnet on the center of the Hall-effect chip as shown in Fig. \ref{fig:sens_setup}.
\par The process consists of moving the robot end effector along each axis and collecting magnetic field readings at a total of thirty discrete points. To determine the mapping in the Z-axis, the robot moved the magnet away in 0.2mm increments withing the range $[1,3]$mm of distance between the magnet and the chip. Similarly, in the X and Y dimensions, the data is collected in the range $[-1,1]$mm with 0.2mm increments in their respective axes.
% The radial component of the data collection process is reversed in terms of mounting for the chip and the magnet. The chip is mounted on the cartesian robot for alignment purpose, and the magnet is mounted on a rotary stage as shown in figure \ref{fig:sens_setup}. The data is collected about the X and Y axes within three radiuses (1, 1.5 and 2 mm). On every radius a total of 5 readings were collected spanning -5 deg to +5 deg in 2.5 deg increments. Which sums up to a total of 15 points collected about each axis.

\par
\begin{figure}[t]
     \centering
     % trim={<left> <lower> <right> <upper>}
     \includegraphics[width=1.0\columnwidth]{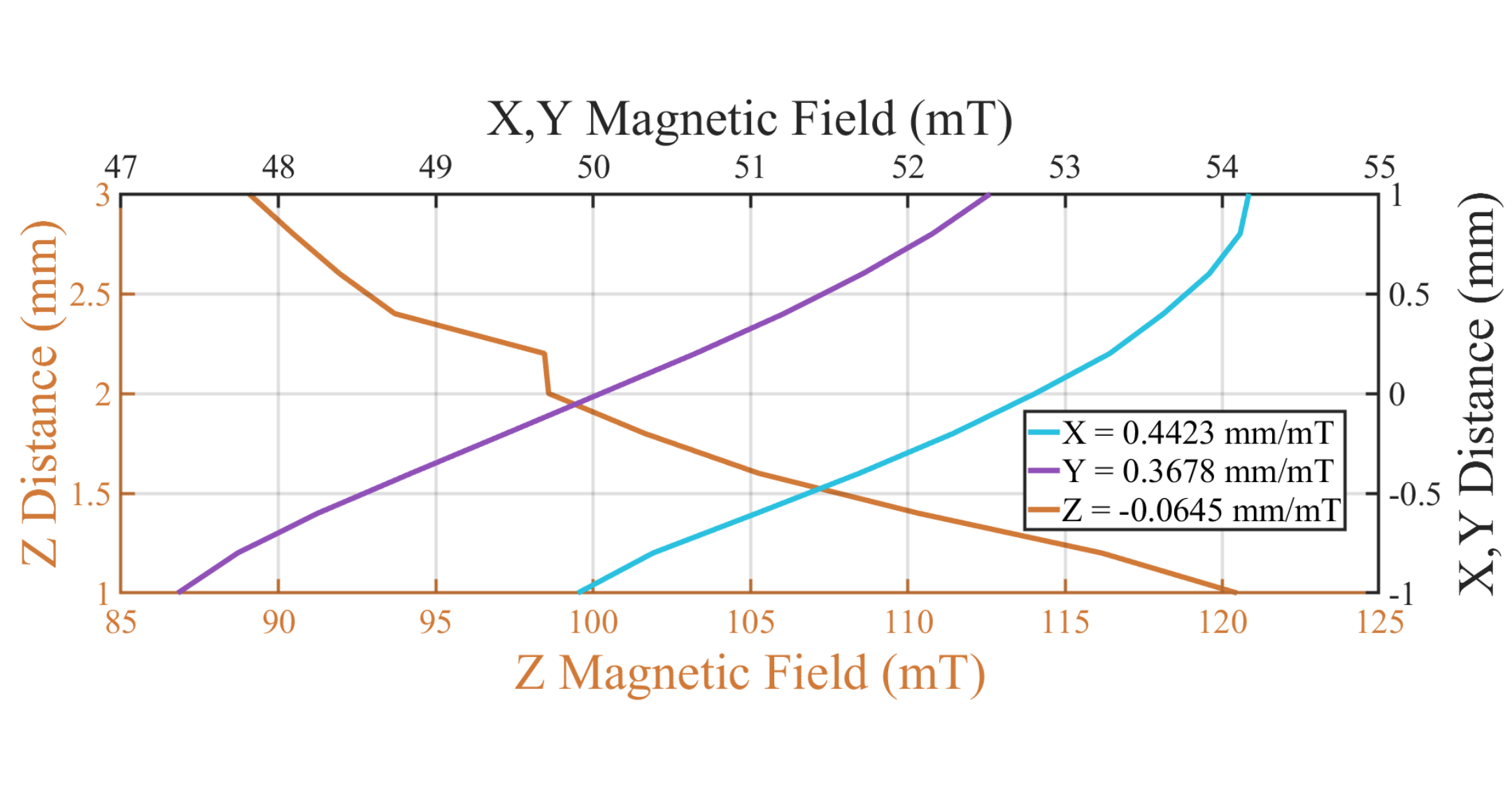}
        \caption{Distance of magnet from sensor frame as a function of magnetic field strength for X, Y, and Z axes. Note that although the range of distances and magnetic flux vary between X/Y and Z, the plots are superimposed. }
        \label{fig:sensitivity}
\end{figure}

 The resulting mapping between position and magnetic flux density along each of the three axes is shown in Fig. \ref{fig:sensitivity}. The results demonstrate that the relationship between magnetic field strength and position is approximately linear in each dimension \corr{for the anticipated range}. Therefore, \eqref{eq:map} is can be written with $\mathbf{M}$ populated by the slopes and $\mathbf{o}$ from the y-intercepts from the linear regression model of each curve:

\begin{equation}
    \label{eq:M_matrix}
    \mathbf{p} = \underbrace{\begin{bmatrix}
        0.4423 & 0 & 0\\
        0 & 0.3678 & 0\\
        0 & 0 & -0.0645
    \end{bmatrix}}_{\mb{M}} \mathbf{b} + \underbrace{\begin{bmatrix}
        -22\\
        -18\\
        8
    \end{bmatrix}}_{\mb{o}}
\end{equation}

% The offset, $\mathbf{o}$, is unique to each Hall Effect sensor and is computed by generating a baseline reading from the force sensor when no force is applied. From this, the magnetic field strength detected by each magnetometer is multiplied by $\mathbf{M}$ and subtracted from the nominal distance between the chips and the magnets (i.e. $\mathbf{p} = [0, 0, 6]^T$). 

\subsection{Calibration}
\label{sec:calibration}

\par
\begin{figure*}[h]
    \centering
    % trim={<left> <lower> <right> <upper>}
     \includegraphics[angle=0, origin=c,trim={0.0cm 5cm 0.5cm 0cm},clip,width=1.0\textwidth]
    {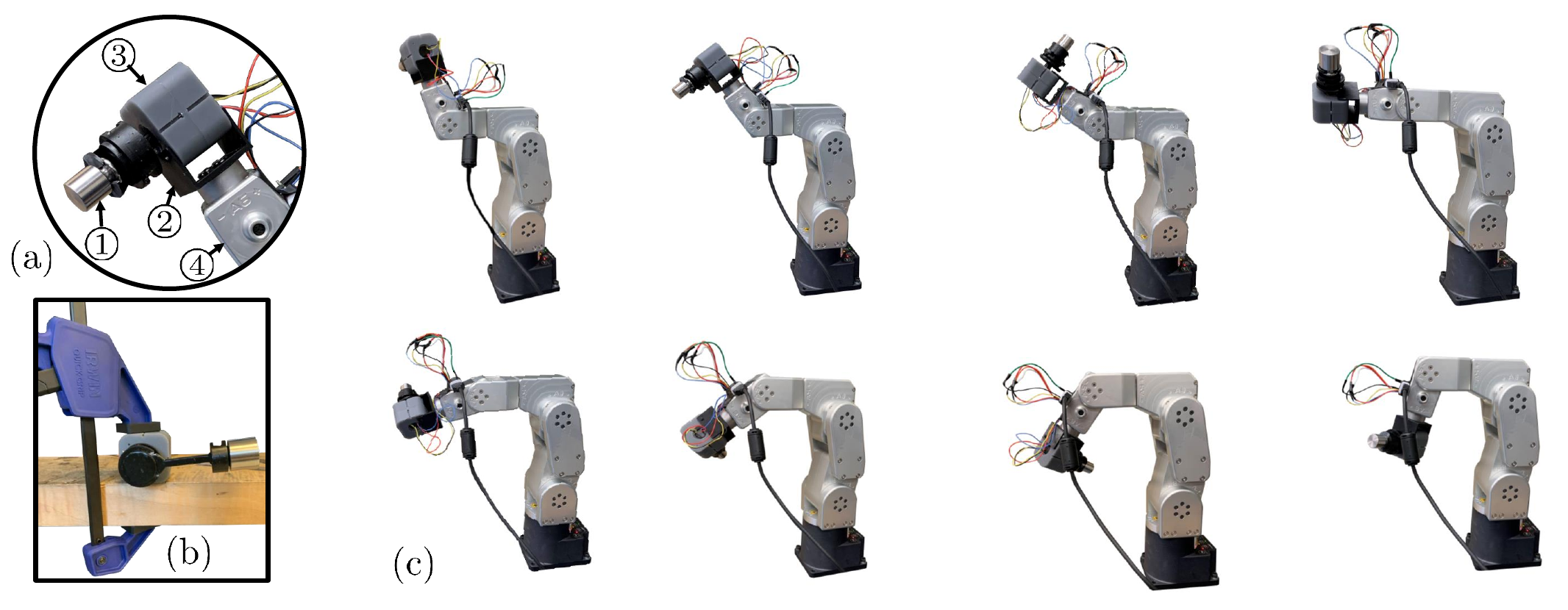}
    \caption{Calibration experimental setup. (a) Closeup of force sensor attached to calibration weight and robot flange. \protect\circled{1} 100g mass. \protect\circled{2} 3D printed attachment. \protect\circled{3} force sensor. \protect\circled{4} Meca500 robot (b) Experimental setup for z axis torque experiments. (c) Example eight poses out of 193 used to calibrate the force sensor. Multimedia Extension I shows this procedure. }
    \label{fig:calib_setup}
\end{figure*}

 The goal of the calibration procedure is to construct the stiffness matrix, $\mathbf{K}$, in order to relate the magnetic flux readings to force \eqref{eq:wrench}. The outer shell of the force sensor is fixed to the end effector of the Meca500 (Mecademic) robot. A rod which has an integrated attachment for calibration weights is rigidly connected to the inner ring of the force sensor. Masses of 50g and 200g were used to calibrate the sensor. The robot was commanded to put the sensor in 193 different poses to sample a robust combination of forces and torques across all six axes (Fig. \ref{fig:calib_setup}). The mean of a hundred measurements from each Hall-effect sensor were taken at each pose to reduce the effect of noise in the measurements. A video of these experiments is provided in Multimedia Extension I.

\par From the Hall-effect sensor readings and the method described in section \ref{sec:center_pose_est} and \ref{sec:deflection_twist}, the deflection twist of the sensor frame is computed for each pose. These twists make up the $\bs{\Xi}$ matrix which is substituted into \eqref{eq:solve_A} with the Hall-effect readings to generate $\mathbf{A}$, which estimates the deflection twist directly from the set of magnetic flux measurements. 

\par To determine the wrenches applied to the force sensor at each pose, they must first be defined in world frame. The position of the center of mass of the calibration weight is given by the direct kinematics of the robot and a static transform between the end effector frame \{$ee$\} and the center of mass frame \{$\omega$\}, which is measured from CAD \eqref{eq:p}. 

\begin{equation}\label{eq:p}
    \begin{bmatrix}
        ^{0}\mb{p}_{\omega} \\ 1
    \end{bmatrix} = \htf{0}{ee}\begin{bmatrix}
         {}^{ee} \mb{p}_{\omega} \\ 1
    \end{bmatrix}
\end{equation}

The orientation of the \{$\omega$\} frame is aligned with the world frame so that the force applied at \{$\omega$\} is trivially $\mathbf{w_\omega} = [0, 0, -mg, 0, 0, 0]^T$. The resulting force felt by the force sensor is calculated by taking multiplying the $\omega$ force by an adjoint transpose:

\begin{equation}
    \mathbf{w}_c = \mb{Ad}\T \mathbf{w_\omega}
\end{equation}
Where the adjoint matrix is given by \eqref{eq:adj} for the rotation and position between the sensor frame and the calibration weight frame, $\Rot{\omega}{C}$ and $^{\omega}\mathbf{p}_C$, respectively. 

\begin{equation} \label{eq:adj}
    \mb{Ad} = \begin{bmatrix}
         \Rot{\omega}{c} & ^{\omega}\mathbf{p}_c {}^\wedge \Rot{\omega}{c} \\
         \mathbf{0} & \Rot{\omega}{c}
    \end{bmatrix}
\end{equation}

Now that the deflection twist and the wrench on the sensor is known for each pose, the stiffness matrix can be computed from equation \eqref{eq:solve_stiffness}:
\begin{equation}
    \label{eq:K_matrix}
    \scalebox{0.85}{$\mathbf{K} = \begin{bmatrix}
        -8.82  &  -11.33  &  -17.41  &  0.03 &  0.14 &  0.02\\
    1.53  & 12.77  &  0.65 &  -0.21 &   -0.02 &  -0.06\\
    39.70 &   27.12 &  -6.50 &  -0.03  &  0.10  &  -0.04\\
   -0.13  &  -0.43 &  -0.02 &  0.01 &  0.00  &  0.00\\
   -0.76  &  -0.54  &  -0.92  &  0.00  &0.01  & 0.00\\
    0.08  &  0.02 &  0.03  & 0.00  &  0.00  &  0.01\\
    \end{bmatrix} \times 10^3$}
\end{equation} 
 To report the upper bound of error propagation from the Hall-effect measurements, the singular values of $\mb{KA}$ are investigated. For forces, $\sigma_{\max} = 6.07\times 10^{-3}N/\mu T$, $\sigma_{\min} = 2.88\times 10^{-3}N/\mu T$ and the isotropy index is $\sigma_{\min}/\sigma_{\max} = 0.47$. As for the torques, $\sigma_{\max} = 2.26\times 10^{-3}Nm/\mu T$, $\sigma_{\min} = 1.48\times 10^{-3}Nm/\mu T$ and the isotropy index is $\sigma_{\min}/\sigma_{\max} = 0.65$.

% \begin{figure}[h]
%     \centering
%     \includegraphics[width=0.5\columnwidth]{figures/pdf/ellipsoids.pdf}
%     \caption{ Force sensing sensitivity ellipsoids: (a) Force, (b) moment\red{Random values for now --> for actual figure, place ellipsoids side by side to save space} }
%     \label{fig:ellipsoids}
% \end{figure}

\subsection{Validation} \label{sec:validation}
\begin{figure*}[h]
    \centering
    % trim={<left> <lower> <right> <upper>}
    % [angle=0, origin=c,trim={0.0cm 5cm 0.5cm 0cm},clip,width=1.0\textwidth]
     \includegraphics[width=0.9\textwidth]{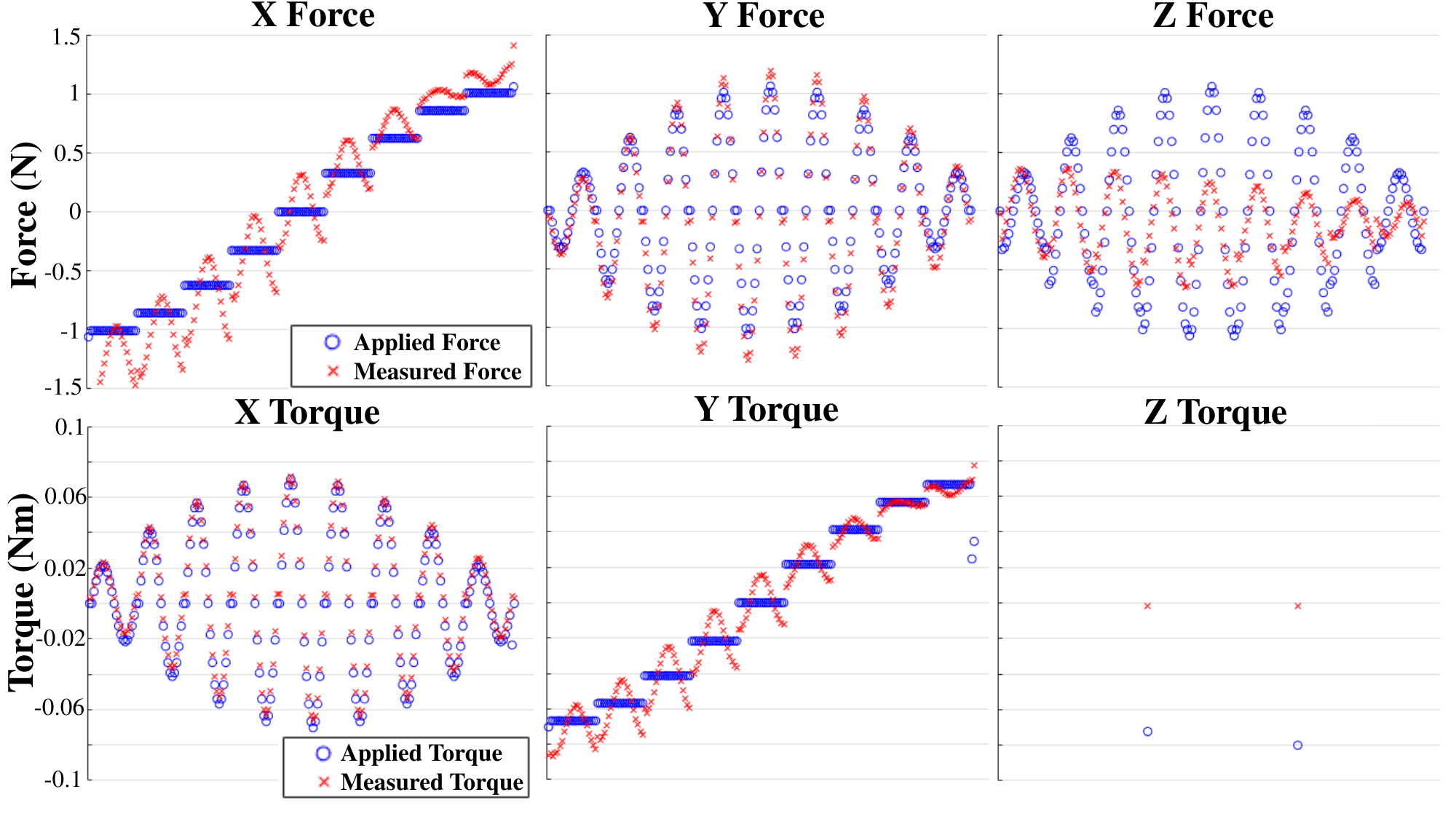}
    \caption{Validation results of the force sensor that compares the applied and measured force/torque in each dimension. }
    \label{fig:wrench_plots}
\end{figure*}

% \begin{figure}[h]
%     \centering
%     % trim={<left> <lower> <right> <upper>}
%     % [angle=0, origin=c,trim={0.0cm 5cm 0.5cm 0cm},clip,width=1.0\textwidth]
%      \includegraphics[angle=0, origin=c,trim={2cm 20cm 4cm 0cm},clip,width=\columnwidth]{figures/pdf/vertical_wrench_plots.pdf}
%     \caption{Some stuff}
%     \label{fig:wrench_plots}
% \end{figure}

\par The purpose of the validation experiment is to verify the accuracy of the force sensor by comparing a known applied force to the measured force. The setup was identical to that used for calibration, except a different calibration weight was used, specifically, 100g was chosen. As in the calibration procedure, the sensor is held in different poses and the magnetic flux density from each Hall Effect sensor is recorded. Using the method from \ref{sec:center_pose_est} and \ref{sec:deflection_twist}, the deflection twist was generated from the data, then the measured force by the sensor was computed using \eqref{eq:wrench}. The ground truth force/torque values applied to the center were calculated according to the method previously described in section \ref{sec:calibration}. The measured force was compared to the ground truth \corrlab{R4-1}{as shown in Fig. \ref{fig:wrench_plots}.} The root mean square error (RMSE) over all sampled poses is reported in Table \ref{tab:results}. 

\begin{table}[H]
    \caption{Root mean square error of force and torque measurements for each axis.}
    \label{tab:results}
    \centering
    \begin{tabular}{|c|c|c|c|c|c|c|}
    \hline
        \textbf{Quantity} & $F_x$ & $F_y$ & $F_z$ & $M_x$ & $M_y$ & $M_z$\\
        \hline
        \textbf{RMS error} & 0.2338 & 0.1136 & 0.3775 & 4.1 & 10.6 & 8.5\\
        \hline
        \textbf{Units}& N & N & N & mNm& mNm& mNm\\
        \hline
    
    \end{tabular}
\end{table}

\par An overview of the sensor characteristics is provided here. Taking the Euclidean norm of the reported RMS errors for force and torque can summarize the overall error as 0.45N and 0.014Nm, respectively. The final calculated cost of a single force sensor is \$32.74 based on the calculations in Table \ref{tab:cost}. Though not validated experimentally, the estimated range of our sensor $\pm$50N in the X,Y directions, $\pm$20N in the Z directions, and $\pm$0.2Nm for all torques. The estimated range of the force sensor is computed by multiplying the stiffness for a single dimension (elements on the diagonal of $\mathbf{K}$) by the maximum possible deformation in that dimension. \corrlab{R7-5}{For the X and Y directions, the maximum possible deformation is 6mm and in the Z direction it is 3mm.}
\section{Discussion}\label{sec:discussion}
\par Our results indicate that a soft, low-cost, 6-axis force/torque sensor for haptic feedback during soft-tissue surgery is feasible. However, there are a number of design improvements that can be made to increase the accuracy and sensitivity of the F/T sensor. First, the assumption of a linear relationship between magnetic flux density and distance to the magnet likely impedes the accuracy of our force sensing model. Future investigation into a non-linear mapping based on the magnetic dipole model should be considered. In turn, this will reduce the error and improve the sensitivity of the device. \corrlab{linearity assumption}{The linear assumption of silicone deformation must also be revisited to accommodate greater forces and improve the accuracy of the device.} \corrlab{R7}{Once these two assumptions have been relaxed, a thorough investigation into other properties of the sensor such as hysteresis, drift, and repeatability can be conducted.}

\section{Conclusion}\label{sec:conclusion}
This paper presented a low-cost 6-axis load cell designed specifically for haptic surgical trainers or future use in surgical applications. At the outset, our motivation was to design a low-cost, standalone 6-axis F/T sensor that does not require expensive external signal conditioning and amplification. We presented an exploratory design of the sensor using a soft elastomeric matrix, embedded magnets, and moving Hall-effect sensors. The design requires a single 4-line connection for I2C communication and it was shown to achieve a good accuracy on the first attempt of design parameter selection based on an educated guess for material selection and geometry.     
Both the physical design and the theoretical force sensing model were discussed, and a model for force sensing at the tip of a surgical instrument passing through two F/T sensors was introduced. This exploratory study has limitations in that the process of sensor design was \textit{ad-hoc} and the process of geometric and material parameter optimization is a problem that will require substantial future effort. Furthermore, for sensors of this type to work for a large full-scale limit would require hyper-elastic constitutive modeling to be incorporated within the design optimization process. We expect this type of sensor to be useful for low-cost or single use applications and potentially to support the deployment of low-cost technologies in low-resource settings.

%----------------------------------------------------------------------------------------------------------%
%----------------------------------------------REFERENCES--------------------------------------------------%
%----------------------------------------------------------------------------------------------------------%
% \newpage
\bibliographystyle{IEEEtran}
\bibliography{bib/madison_bib,bib/nabil_bib,bib/Garrison_bib}

\balance
\end{document}